\documentclass[letterpaper, 10 pt, conference]{ieeeconf}  

\IEEEoverridecommandlockouts                              

\newcommand{\PaperVersion}{arxiv}

\newcommand{\youtubeurl}{https://youtu.be/WKF82YBOH-Q}

\usepackage{graphicx}
\usepackage{siunitx}
\usepackage{bm}
\usepackage{amsmath} 
\usepackage{amssymb}  
\usepackage{cite}
\usepackage{xcolor}
\usepackage{ifthen}

\newcommand{\IEEEarxivnotice}{%
  \AddToHookNext{shipout/foreground}{%
    \put(0,-0.35in){%
      \makebox[\paperwidth][c]{%
        \setlength{\fboxsep}{2pt}%
        \fbox{%
          \begin{minipage}{0.98\textwidth}%
            \footnotesize
            \setlength{\parskip}{0pt}%
            \setlength{\baselineskip}{9.5pt}%
            \textbf{Accepted for publication in IEEE Conference on Decision and Control 2026.}\ %
            \textcopyright\ 2026 IEEE. Personal use of this material is permitted. Permission from IEEE must be obtained for all other uses, in any current or future media, including reprinting/republishing this material for advertising or promotional purposes, creating new collective works, for resale or redistribution to servers or lists, or reuse of any copyrighted component of this work in other works.%
          \end{minipage}%
        }%
      }%
    }%
  }%
}
\usepackage[
	pdftitle={Accurate Open-Loop Control of a Soft Continuum Robot Using Visually Learned Latent Dynamics},
	pdfsubject={IEEE Conference on Decision and Control (CDC 2026)},
	pdfauthor={Henrik Krauss, Johann Licher, Naoya Takeishi, Annika Raatz, Takehisa Yairi},
	pdfkeywords={soft robotics, latent dynamics, open-loop control, visual oscillator networks},
    hidelinks
]{hyperref}
\usepackage{url}
\usepackage{subcaption}

\ifthenelse{\equal{\PaperVersion}{diff}}{%
  \newcommand{\revcolor}{red}%
}{%
  \newcommand{\revcolor}{black}%
}
\newcommand{\rev}[1]{{\color{\revcolor}#1}}

\DeclareMathOperator*{\argmin}{arg\,min}

\title{\LARGE \bf
Accurate Open-Loop Control of a Soft Continuum Robot\\
Using Visually Learned Latent Dynamics
}

\author{Henrik Krauss$^{1}$, Johann Licher$^{2}$, Naoya Takeishi$^{3}$, Annika Raatz$^{2}$, Takehisa Yairi$^{3}$ 
\\
\thanks{This work was supported in part by JST PRESTO under Grant JPMJPR24T6, in part by JSPS KAKENHI under Grant JP25H01454, and in part by the Deutsche Forschungsgemeinschaft (DFG, German Research Foundation) under Grant No. 405030609.}
\thanks{$^{1}$Henrik Krauss is with Department of Advanced Interdisciplinary Studies, The University of
Tokyo, Tokyo, 153-8904, Japan.
        {\tt\small henrik1.krauss@gmail.com}}%
\thanks{$^{2}$Johann Licher and Annika Raatz are with the Institute of Assembly Technology and Robotics, Leibniz University Hannover, 30823 Germany.
        {\tt\small licher@match.uni-hannover.de}}
\thanks{$^{3}$Naoya Takeishi and Takehisa Yairi are with the Research Center for Advanced Science and Technology, The University
of Tokyo, Tokyo, 153-8904, Japan.}
}

\begin{document}

\thispagestyle{empty}
\pagestyle{empty}

\maketitle
\ifthenelse{\equal{\PaperVersion}{arxiv}}{\IEEEarxivnotice}{}%
\thispagestyle{empty}
\pagestyle{empty}

\begin{abstract}
This work addresses open-loop control of a soft continuum robot (SCR) from video-learned latent dynamics. Visual Oscillator Networks (VONs) from previous work are used, which provide mechanically interpretable 2D oscillator latents through an attention broadcast decoder (ABCD). Open-loop, single-shooting optimal control is performed in latent space to track image-specified waypoints without camera feedback. An interactive SCR live simulator enables design of static, dynamic, and extrapolated targets and maps them to model-specific latent waypoints. On a two-segment pneumatic SCR, Koopman, MLP, and oscillator dynamics, each with and without ABCD, are evaluated on setpoint and dynamic trajectories. ABCD-based models consistently reduce image-space tracking error. The VON and ABCD-based Koopman models attain the lowest mean squared errors (MSEs). Ablation and simulation stress tests confirm that training and architecture choices support open-loop performance, including static holding, stable extrapolated equilibria, and relaxation to rest. To the best of our knowledge, this is the first demonstration of reliable long-horizon open-loop control of a physical soft robot using interpretable latent dynamics models learned solely from video.
\end{abstract}

\section{Introduction}
\begin{figure}[t]
    \centering
    \includegraphics[width=0.9\linewidth,trim=0 35 0 0]{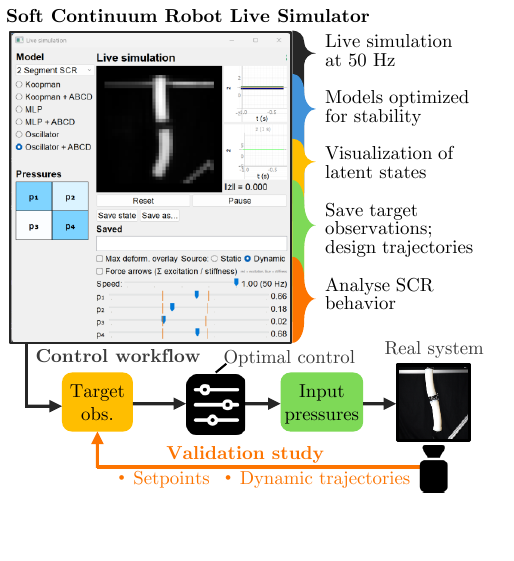}
    \caption{Overview of this study's approach: an SCR live simulator is employed to generate target states for open-loop optimal control of the physical SCR. The feasibility of this method is validated across diverse trajectories.}
    \label{fig:intro}
\end{figure}

Recent developments in data-driven methods have enabled discovery of latent dynamics in high-dimensional data, such as video. Prominent methods are Koopman theory, (extended) dynamic mode decomposition (DMD) and spectral submanifolds (SSMs)~\cite{takeishi2017learning,lusch2018deep,brunton2022modern,cenedese_datadriven_2022}.

Soft continuum robots (SCR) exhibit continuous deformation and a theoretically infinite degree of freedom and are therefore challenging to model and control~\cite{thuruthel_control_2018}.
Classical SCR modelling approaches, such as piece-wise constant curvature (PCC) or Cosserat rod-based dynamics, require manual model derivation and are either limited in accuracy or computational efficiency~\cite{armanini2023soft}. Hybrid approaches have been proposed to overcome these limitations~\cite{falotico_learning_2025}. Licher et al. have achieved highly accurate closed-loop dynamic control of an SCR based on nonlinear evolutionary model-predictive control (MPC) applied to a physics-informed neural network (PINN) based on Cosserat rod dynamics~\cite{licher2025adaptive}.

In contrast to enabling fast solution of computationally expensive, manually derived models, recent developments also focused on deriving low-order representations that describe the complex nonlinear dynamics of SCRs. Examples include low-dimensional strain parameterizations~\cite{alkayas2025structure}, or low-order finite-element models which have been applied to optimal control~\cite{tonkens2021soft}. Furthermore, Koopman theory and SSMs have been used for accurate closed-loop control of SCRs~\cite{haggerty2023control,alora_discovering_2025a}.
As part of fully data-driven \rev{approaches}, Stölzle et al. have introduced coupled oscillator networks (CONs) that learn latent models of soft robots from video, and applied them to control of a simulated SCR~\cite{stoelzle2024input}.

While easily derived from data, they lack mechanical interpretability and evidence of generalization. In our previous work we presented visual oscillator networks (VONs), coupling ABCD attention maps with 2D oscillator positions for on-image interpretable dynamics~\cite{krauss2025learning} that generalize well and show good prediction performance.

However, latent dynamical models are \emph{rarely used open-loop} because prediction errors accumulate over time and can drive the trajectory off the manifold on which the real system dynamics are represented. Some studies have demonstrated open-loop latent dynamical controllers for soft actuators, e.g. by learning inverse models from actuator responses and iterative learning control data~\cite{sugiyama2024latent} or by training a recurrent neural network to predict images from actuation inputs and using it to optimize inputs for trajectory tracking~\cite{marquesmonteiro_visuodynamic_2024}. 

\rev{Open-loop control of SCRs with video-learned recurrent models has been shown, but typically relies on image-level optimization, with limited accuracy and little evidence of structural stability. Building on the VON/ABCD architectures of~\cite{krauss2025learning}, we study whether explicit latent dynamics can support long-horizon open-loop tracking of target images on hardware without camera feedback.} \rev{Targets may also need to come from unseen images, e.g.\ by drawing~\cite{almanzor_static_2023,marquesmonteiro_visuodynamic_2024} or simulation.}

This work addresses three key challenges in learning-based control of SCRs: (i) the lack of interpretability of learned representations for control, (ii) reliable open-loop control without feedback, and (iii) the availability of target states, including unseen or extrapolated configurations. Our main contributions are:
\begin{enumerate}
    \item \rev{We apply VON/ABCD models~\cite{krauss2025learning} to open-loop optimal control of an SCR from visual observations, without a hand-derived dynamics model. We evaluate them on hardware against Koopman and multilayer perceptron (MLP) baselines (Fig.~\ref{fig:intro}), analyze limitations in simulation, and give training guidelines via an ablation (multi-step and rest-state losses, and implicit damping).}
    \item We introduce an SCR live simulator for interactive target state and trajectory design, supporting control with unseen or extrapolated targets (Fig.~\ref{fig:intro}).
\end{enumerate}

\section{Methods}
\subsection{Learning Latent Dynamics from Video}
We learn latent dynamical models from video using an encoder--dynamics--decoder model based on~\cite{lusch2018deep,stoelzle2024input}. For an image $\bm{o}^{(i)}$ at step $i$, the encoder $\varphi$ produces latent coordinates
\begin{equation}
    \bm{z}^{(i)} = \varphi(\bm{o}^{(i)}),
\end{equation}
and the dynamics model $f_{\mathrm{dyn}}$ predicts the next latent coordinates and velocities under input $\bm{u}^{(i)}$,
\begin{equation}
    (\hat{\bm{z}}^{(i+1)}, \hat{\bm{\dot{z}}}^{(i+1)}) = f_{\mathrm{dyn}}(\bm{z}^{(i)}, \bm{\dot{z}}^{(i)}, \bm{u}^{(i)}),
\end{equation}
which is decoded to the next image $\hat{\bm{o}}^{(i+1)} = \varphi^{-1}(\hat{\bm{z}}^{(i+1)})$.

Following~\cite{stoelzle2024input}, we define the full latent state as $\bm{\xi}^{(i)} = [\bm{z}^{(i)\top}, \bm{\dot{z}}^{(i)\top}]^\top$. The latent velocity is obtained from the encoder Jacobian,
\begin{equation}
    \dot{\bm{z}}^{(i)} =
    \frac{\partial \varphi}{\partial \bm{o}}(\bm{o}^{(i)}) \, \dot{\bm{o}}^{(i)},
    \label{eq:jacobian}
\end{equation}
and central finite differences of observations.
We use a $\beta$-\rev{variational autoencoder (VAE)}~\cite{higgins2017beta} and train end-to-end with static and dynamic image reconstruction, and latent consistency losses, originally based on~\cite{lusch2018deep}, and with \rev{architecture details following~\cite{krauss2025learning}}.

Three latent dynamical models are compared, where for all models $\bm{B}(\cdot)$ is parameterized by an MLP:

\textit{(I) Koopman:} We use a transition matrix $\bm{A}$ for the latent-state update
\begin{equation}
    \bm{\xi}^{(i+1)} = \bm{A}\bm{\xi}^{(i)} + \bm{B}(\bm{u}^{(i)}).
\end{equation}

\textit{(II) MLP:} Providing a flexible baseline, an MLP predicts the next latent velocity as
\begin{equation}
    \bm{\dot{z}}^{(i+1)} = f_{\mathrm{MLP}}(\bm{\xi}^{(i)}) + \bm{B}(\bm{u}^{(i)}),
\end{equation}
from which the next latent coordinate
\begin{equation}
    \bm{z}^{(i+1)} = \bm{z}^{(i)} + \Delta t \, \bm{\dot{z}}^{(i+1)}
\end{equation}
is obtained from integration to ensure kinematic consistency.

\textit{(III) Oscillator network:} The latent coordinates follow the equation of motion
\begin{equation}
    \bm{M}\ddot{\bm{z}} + \bm{D}\dot{\bm{z}} + \bm{K}(\bm{z} - \bm{z}_0) = \bm{B}(\bm{u}),
\end{equation}
where $\bm{z}_0$ is an optional, learnable non-zero rest position used by VONs, and $\bm{M}$, $\bm{D}$, $\bm{K}$ denote mass, damping, and stiffness matrices. We integrate this system using a symplectic Euler scheme with implicit damping:
\begin{align}
    \bm{\dot{z}}^{(i+1)} &= \bm{\Gamma}^{-1}\!\bigl[\bm{\dot{z}}^{(i)}+\Delta t\,\bm{M}^{-1}\!\bigl(\bm{B}(\bm{u}^{(i)})-\bm{K}(\bm{z}^{(i)}-\bm{z}_0)\bigr)\bigr]\!, \\
    \bm{z}^{(i+1)} &= \bm{z}^{(i)} + \Delta t \, \bm{\dot{z}}^{(i+1)},
\end{align}
with $\bm{\Gamma} = \operatorname{diag}\!\left(\bm{I} + \Delta t \bm{M}^{-1}\bm{D}\right)$.
\rev{Positive-definite $\bm{K},\bm{D}$ and a positive diagonal $\bm{M}^{-1}$ are enforced by construction, yielding an input-to-state stable (ISS) system with respect to the external forcing $\bm{B}(\bm{u})$~\cite{stoelzle2024input}.}

Apart from the newly introduced implicit damping, the Koopman and oscillator formulations are consistent with our previous work. All three models, including the MLP, can be combined with the ABCD introduced in~\cite{krauss2025learning}. In particular, the oscillator model with the attention coupling loss, aligning the relative movements of latent 2D oscillators with corresponding ABCD attention maps, yields the VON.

To improve suitability for open-loop control and extending on~\cite{krauss2025learning}, we further replace the dynamic single-step losses by multi-step losses over an $H$-step rollout, where $H$ increases over training epochs, for the latent dynamical consistency loss
\begin{equation}
\mathcal{L}_\text{d}^{(H)} =
\frac{1}{N}\sum_{n=1}^{N} \frac{1}{H}\sum_{h=1}^{H}
\text{MSE}(\varphi^{-1}(\hat{\bm{z}}^{(n,h)}), \bm{o}^{(n,h)}),
\end{equation}
as well as the dynamic reconstruction loss
\begin{equation}
\begin{aligned}
\mathcal{L}_z^{(H)} =
\frac{1}{N}\sum_{n=1}^{N} \frac{1}{H}\sum_{h=1}^{H}
\Big(
\text{MSE}(\hat{\bm{z}}^{(n,h)}, \bm{z}^{(n,h)}) \\
\qquad + \text{MSE}(\Delta t \cdot \hat{\bm{\dot{z}}}^{(n,h)}, \Delta t \cdot \bm{\dot{z}}^{(n,h)})
\Big).
\end{aligned}
\end{equation}
Here $(n,h)$ denotes the $h$-th step of the $n$-th sequence in a batch of size $N$, and $H$ is the scheduled horizon. \rev{The $H$-step rollout more strongly penalizes open-loop drift and therefore aids keeping long-horizon predictions on-manifold.} For all model types, we additionally use an adjusted, simplified rest-state loss (instead of a steady-state loss)
\begin{equation}
\begin{aligned}
\mathcal{L}_\text{s} =
\frac{1}{2}\Big(
&\text{MSE}(\varphi(\bm{o}_\text{rest}), \bm{z}_0) \\
&+ \Big(
\text{MSE}\!\left(\left[f_\text{dyn}(\varphi(\bm{o}_\text{rest}), \bm{0}, \bm{u}_\text{rest})\right]_1, \bm{z}_0\right) \\
&\qquad + \text{MSE}\!\left(\Delta t \cdot \left[f_\text{dyn}(\varphi(\bm{o}_\text{rest}), \bm{0}, \bm{u}_\text{rest})\right]_2, \bm{0}\right)
\Big)
\Big),
\end{aligned}
\end{equation}
enforcing that rest-state $\bm{o}_\text{rest}$ encodes to $\bm{z}_0$ and stays at equilibrium under rest actuation $\bm{u}_\text{rest}$. The rest state is also the starting point for control from rest.
Finally, for VONs we apply mean correction in the KL term,
\begin{equation}
\mathcal{L}_\text{KL}^{(z_0)} =
\frac{-1}{2N}\sum_{i=1}^{N}\sum_j \Big(1 + \log (\sigma_j^{(i)})^2 - (\mu_j^{(i)} - z_{0,j})^2 - (\sigma_j^{(i)})^2 \Big),
\end{equation}
\rev{where $\mu_j^{(i)}$ and $\sigma_j^{(i)}$ are the encoder VAE mean and standard deviation for latent $j$, and mean-correction centers the Kullback--Leibler (KL) term at the learned rest $z_0$ so rest-state latents are not pushed to~$0$.}

\subsection{Predictive Control in Latent Space}
We use a discrete-time formulation with step index $i=0,\dots,T-1$ at \SI{50}{Hz}. Given an initial latent state $(\bm{z}^{(0)},\bm{\dot{z}}^{(0)})$ and a horizon of $T$ steps, we solve a single-shooting open-loop optimal control problem over the control sequence $\bm{u}^{(0)},\dots,\bm{u}^{(T-1)}$ subject to the learned latent dynamics
\begin{equation}
    (\bm{z}^{(i+1)}, \bm{\dot{z}}^{(i+1)}) = f_\text{dyn}(\bm{z}^{(i)}, \bm{\dot{z}}^{(i)}, \bm{u}^{(i)}), \qquad i=0,\dots,T-1,
\end{equation}
with $(\bm{z}^{(T)},\bm{\dot{z}}^{(T)})$ being the terminal state.

Targets are one or more observations $\{\bm{o}_k^\ast\}_{k=1}^{K}$. Latent targets are $\bm{z}_k^\ast = \varphi(\bm{o}_k^\ast)$; $\bm{\dot{z}}_k^\ast$ is from central finite differences via~\eqref{eq:jacobian} when available, else $\bm{0}$ for static and final targets. For $K>1$, the target points are distributed uniformly over the horizon at state indices $\tau_1,\dots,\tau_K$ with $\tau_1=1$, $\tau_K=T$, e.g.\ $\tau_k = 1 + \mathrm{round}((k-1)(T-1)/(K-1))$, and the active target index $k(i)$ is either the next or the closest target from the current state index $i$:
\begin{equation}
k(i) =
\begin{cases}
\min \{k \in \{1,\dots,K\} \mid \tau_k \ge i\}, & \text{(next target)} \\
\argmin_{k \in \{1,\dots,K\}} |i - \tau_k|, & \text{(closest target)}.
\end{cases}
\end{equation}

With $(\bar{\bm{z}}^{(i)},\bar{\bm{\dot{z}}}^{(i)}) = (\bm{z}_{k(i)}^\ast,\bm{\dot{z}}_{k(i)}^\ast)$, we solve the optimal control problem
\begin{equation}
\begin{aligned}
&\min_{\bm{u}^{(0:T-1)}}\!J
= \underbrace{\tfrac{w_Q}{T}\sum_{i=1}^{T}\|\bm{z}^{(i)}-\bar{\bm{z}}^{(i)}\|^2+\tfrac{w_{\dot{Q}}}{T}\sum_{i=1}^{T}\|\bm{\dot{z}}^{(i)}-\bar{\bm{\dot{z}}}^{(i)}\|^2}_{\text{next/closest waypoint}} \\
&\quad + \underbrace{\tfrac{w_{Q_k}}{K}\sum_{k=1}^{K}\|\bm{z}^{(\tau_k)}-\bm{z}_k^\ast\|^2+\tfrac{w_{\dot{Q}_k}}{K}\sum_{k=1}^{K}\|\bm{\dot{z}}^{(\tau_k)}-\bm{\dot{z}}_k^\ast\|^2}_{\text{waypoint-exact cost}} \\
&\quad + \underbrace{w_{Q_f}\|\bm{z}^{(T)}-\bar{\bm{z}}^{(T)}\|^2+w_{\dot{Q}_f}\|\bm{\dot{z}}^{(T)}-\bar{\bm{\dot{z}}}^{(T)}\|^2}_{\text{terminal cost}} \\
&\quad + \tfrac{w_R}{T-1}\sum_{i=1}^{T-1}\|\Delta\bm{u}^{(i)}\|^2+\tfrac{w_{\Delta u}}{T-1}\sum_{i=1}^{T-1}\phi(\Delta\bm{u}^{(i)}),
\end{aligned}
\end{equation}
where the weights $w_\cdot$ trade off waypoint, terminal, and control-rate terms, $\Delta \bm{u}^{(i)} = \bm{u}^{(i)}-\bm{u}^{(i-1)}$, and $w_{\Delta u}$ penalizes only increments exceeding a prescribed bound after
\begin{equation}
        \phi(\Delta\bm{u}) = \|\max(|\Delta\bm{u}|-\Delta\bm{u}_{\max},\bm{0})\|_2^2.
\end{equation}
with absolute value and max applied element-wise, to further respect limits of the SCR pressure controller.

In practice, state and velocity errors in the cost are scaled by the mean per-dimension standard deviation of latents on validation data, so loss weights are comparable across models and relative structure is preserved.
\rev{The problem is solved by single-shooting: $\bm{u}^{(0)}$ is held at the initial static pressures and $\bm{u}^{(1:T-1)}$ are updated by Adam through the latent rollout (typically 200 iterations), with $\Delta\bm{u}$ bounded as above.}

\subsection{Obtaining Target Observations Through an SCR Live Simulator}
\label{subsec:simulator}
We present an SCR live simulator in Fig.~\ref{fig:intro}, built on Python and PyQtGraph, that shows the SCR as predicted by Koopman, MLP, and oscillator models with or without ABCD under user input. Pressure sliders and controls allow interactive design of static, dynamic, and extrapolated targets in image space. Pressures outside the dataset range yield unseen targets and can test extrapolation of the learned models. Saved states store decoded observations and applied input, so target latent states and velocities can be derived per model. \rev{Code including the live simulator is available at}\footnote{Code: \href{https://github.com/UThenrik/visual_oscillators_for_SCR}{\texttt{github.com/UThenrik/visual\_oscillators\_for\_SCR}}}.

\section{Experiments}
\subsection{Hardware, Dataset, and Training of Models}
We use a two-segment SCR with three pressure chambers per segment. Two chambers per segment are pressurized equally, providing four effective inputs and planar motion perpendicular to the camera axis. Further details on the dimensions and manufacturing of the actuator are given in~\cite{bartholdt_parameter_2021}. The dataset is a modified version of~\cite{krauss2025learning}, with two 15-minute recordings at \SI{50}{Hz} (pressures down-sampled to this rate). The first uses smooth sinusoidal excitations and linear interpolation over \SIrange{0}{86}{kPa} (extended to lower pressures than the prior version), the second uses step excitations. Both were captured with a fixed industrial camera for repeatable validation. The dataset is released publicly\footnote{Dataset: \url{https://doi.org/10.5281/zenodo.22790365}}. Models are trained only on the sinusoidal excitations dataset, with the step dataset reserved for validation. We train six main models for control evaluation, namely Koopman, MLP, and Oscillator networks, each in plain form and with the ABCD decoder. The oscillator network with ABCD is the VON. Seven ablation VONs are also trained to assess the impact of architecture and hyperparameter choices on open-loop control performance. Training code and configurations, building on~\cite{krauss2025learning}, are provided in the same code repository as the live simulator.

\subsection{Open-Loop Control}
\label{subsec:open-loop-validation}
We evaluate open-loop control for setpoints and dynamic trajectories. Setpoint tasks test rest-to-target, target-to-target and target-to-rest control for previously tested (normal) and slightly extrapolated states\rev{, each later setpoint continuing from the previous predicted terminal}. Dynamic tasks include tracking multiple waypoints, reaching upswing positions (outside the dataset range), and long-horizon prediction. Target observations are obtained in two ways. \emph{Setpoint Normal} uses frames from the step-validation dataset, encoded per model. All other target observations are designed in the live simulator (Sec.~\ref{subsec:simulator}) using the plain Koopman model, and using the VON for the long dynamic trajectories. Table~\ref{tab:trajectories} summarizes the trajectory types and control parameters.
\begin{table}[htbp]
\centering
\caption{Trajectory types and control parameters.\\$n$~=~number of trajectories tested.}
\label{tab:trajectories}
\setlength{\tabcolsep}{2.5pt}
\begin{tabular}{@{}lrrrrrrrrrr@{}}
\hline
Name & $n$ & $T$ & $K$ & $w_Q$ & $w_{\dot{Q}}$ & $w_{Q_k}$ & $w_{\dot{Q}_k}$ & $w_{Q_f}$ & $w_{\dot{Q}_f}$ \\
\hline
Setp. Normal & 9 & 100 & 2 & 1 & 0.002 & 0 & 0 & 0 & 0 \\
Setp. Extrap. & 6 & 100 & 2 & 1 & 0.002 & 0 & 0 & 0 & 0 \\
Dyn. Normal & 5 & 200 & 8 & 1 & 0 & 0 & 0 & 0 & 0.002 \\
Dyn. Fast & 3 & 150 & 8 & 1 & 0 & 0 & 0 & 0 & 0.002 \\
Dyn. Upswing & 3 & 150 & 3 & 0 & 0 & 1 & 0.002 & 0 & 0 \\
Dyn. Long & 5 & 1000 & 22 & 1 & 0 & 0 & 0 & 0 & 0.002 \\
\hline
\end{tabular}
\end{table}

\begin{figure}[h]
    \centering
    \includegraphics[width=0.9\linewidth, trim= 0 10 0 0]{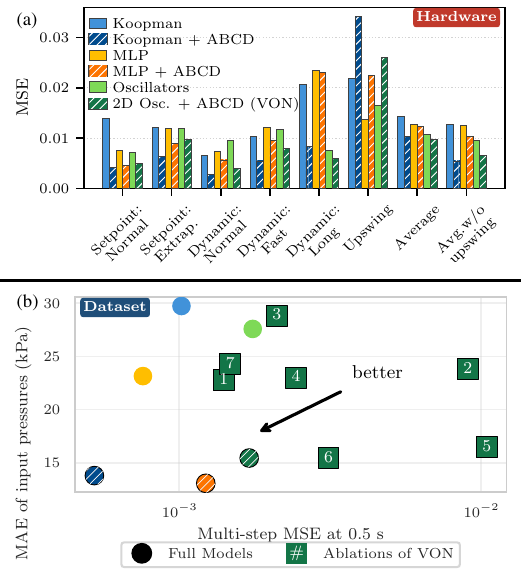}
    \caption{(a) Model performances in open-loop control of real hardware over various trajectory types (image MSEs). (b) Mean absolute error (MAE) of input pressure prediction from open-loop setpoint control in simulation, compared to setpoints in the dataset, over multi-step prediction errors (50 samples each), including ablations.}
    \label{fig:performance}
\end{figure}
\begin{figure*}[h]
    \centering
    \includegraphics[width=0.95\linewidth]{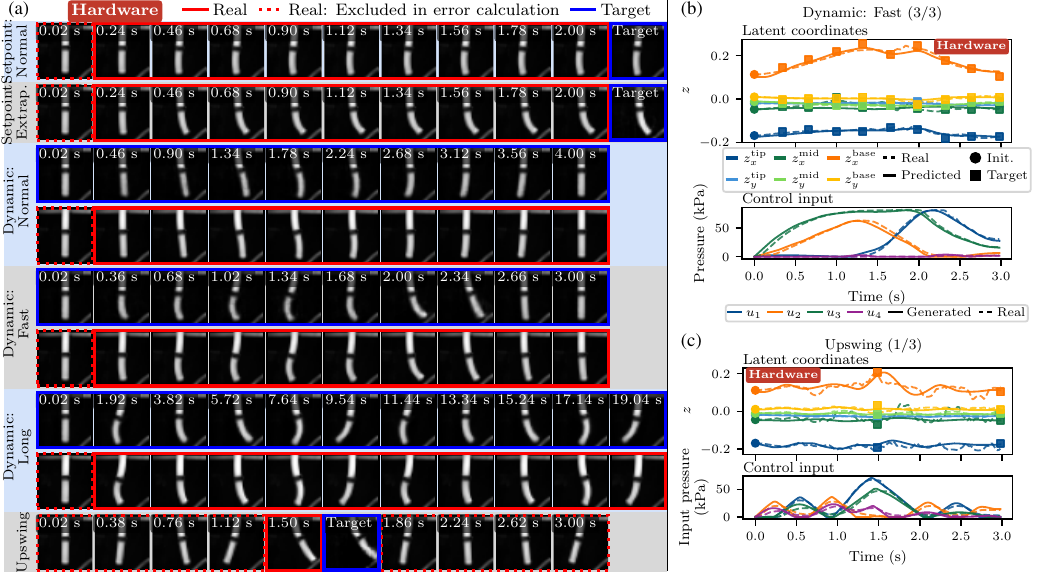}
    \caption{(a) \rev{Hardware} examples for trajectories evaluated for open-loop optimal control. The real trajectories shown are based on the VON model. All target trajectories are accurately achieved, except the upswing, where an additional plot highlights a deviation between set and measured pressure inputs. (b) Predicted and real latent trajectories and input pressures for the fast dynamic trajectory and upswing trajectory (c), depicted on the left.}

    \label{fig:image_results}
\end{figure*}
\rev{Optimized controls (Sec.~\ref{subsec:open-loop-validation}) are executed on the physical SCR (Hardware). Decoded images are $32\times 32$ with pixels in $[0,1]$; MSE is the mean over all pixels and channels,} per trajectory and model over all waypoints except the initial. For setpoints, all post-initial observations are included.

Fig.~\ref{fig:performance}(a) shows the average MSEs for all six models, and Fig.~\ref{fig:image_results}(a) gives representative trajectory examples. Models with ABCD decoders consistently outperform their plain counterparts, except for upswing trajectories. The Koopman and oscillator-based ABCD (i.e., VON) models achieve the lowest overall MSEs at $1.03 \times 10^{-2}$ and $9.80 \times 10^{-3}$. Models are trained on measured pressures, so optimization disregards the dynamics of the low-level pressure controller. \rev{The main hardware limitation is the sim--hardware gap in Fig.~\ref{fig:image_results}(b)--(c) (generated vs.\ measured pressure; predicted vs.\ real latents),} as the low-level pressure controller is too slow under fast upswing oscillations. In contrast, for the faster dynamic trajectories the required pressure changes are within the controller's capabilities and measured pressures match the generated ones well (Fig.~\ref{fig:image_results}(b)).
Fig.~\ref{fig:performance} also gives average MSEs excluding upswing. Using the ABCD consistently reduces open-loop control errors excluding upswing trajectories: for Koopman, the MSE drops from $1.28 \times 10^{-2}$ to $5.45 \times 10^{-3}$; for MLP, from $1.25 \times 10^{-2}$ to $1.03 \times 10^{-2}$; and for the Oscillator, from $9.60 \times 10^{-3}$ to $6.55 \times 10^{-3}$.
\rev{A supplemental video shows hardware runs with camera-versus-target image MSE next to the footage.\footnote{Video: \href{\youtubeurl}{\texttt{https://youtu.be/WKF82YBOH-Q}}}}

\subsection{Ablation study}
\label{subsec:ablation}
We ablate design choices that enable stable open-loop control~\cite{krauss2025learning} to guide video-based latent dynamical modeling for control. We train seven ablation models. In each, exactly one choice is changed from the full VON. The ablation models use: (1)~linear excitation $\bm{B}$ instead of an MLP, (2)~loss weights that prioritize dynamic image reconstruction as in~\cite{krauss2025learning,stoelzle2024input}, (3)~VAE $\beta = 0.01$ instead of $0.0001$, (4)~no rest-state loss, (5)~no multi-step loss, (6)~Rayleigh damping instead of full damping, (7)~symplectic Euler without implicit damping. All other settings in each ablation match the full model.

As an indicator for open-loop setpoint control performance without hardware validation, we use MAE between predicted and real input pressures on 50 setpoints from the step excitation dataset (setpoint optimization as in Sec.~\ref{subsec:open-loop-validation}). Fig.~\ref{fig:performance}(b) plots this pressure MAE against multi-step image MSE on held-out recordings (50 validation trajectories). ABCD-based models yield lower pressure MAE and better multi-step MSE, except for the MLP-based models. The overall best performer is Koopman + ABCD, with a multi-step MSE of $5.23 \times 10^{-4}$ and a pressure MAE of $\SI{13.79}{kPa}$. Within the VON ablation study, the full model lies bottom left, achieving a multi-step MSE of $1.71 \times 10^{-3}$ and the lowest pressure MAE of $\SI{15.44}{kPa}$, compared to ablation averages of $4.92 \times 10^{-3}$ and $\SI{22.08}{kPa}$. We conclude all seven changes likely contribute to accurate open-loop control. Note that pressure MAE can stay high when image MSE is low for visually similar configurations.

\subsection{Simulation experiments}
\begin{figure*}[t]
    \centering
    \includegraphics[width=0.95 \linewidth,keepaspectratio,trim = 0 5 0 0]{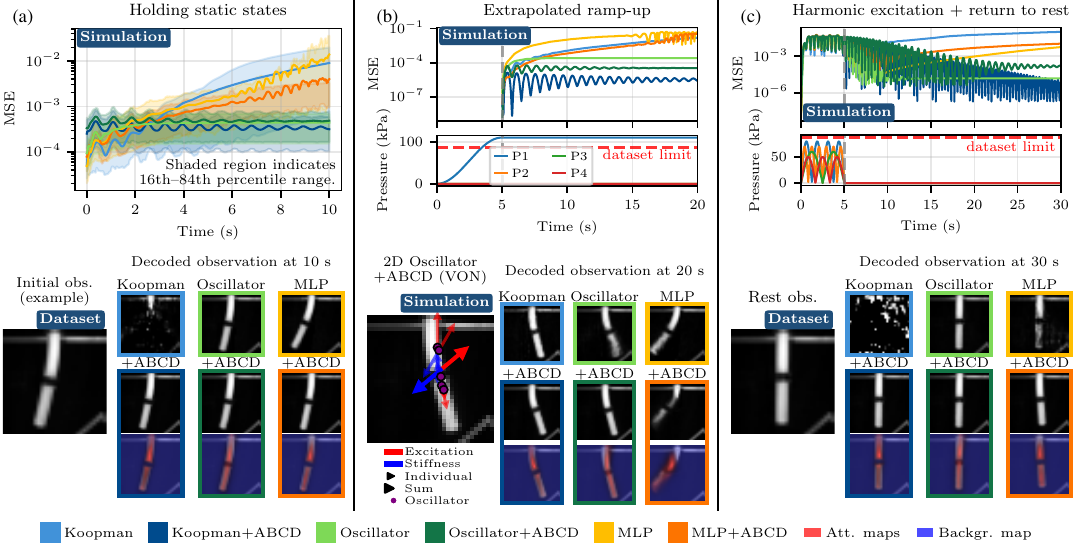}
    \caption{\rev{Simulation stress tests. (a)~Static holding. (b)~Ramp-up to extrapolated pressures, with VON excitation vs.\ stiffness forces. (c)~In-dataset range harmonic excitation then release. Top shows image MSEs and bottom final decoded observations.}}

    \label{fig:sim_results}
\end{figure*}
\rev{Three stress tests carried out in simulation are visualized in~Fig.~\ref{fig:sim_results}. First (a), static holding: 50 static states from the dataset are held at constant pressure for $500$ steps ($\SI{10}{s}$), measuring image MSE to the initial observation. Second (b), a pressure ramp beyond the dataset limit is applied and held, to test whether the decoded pose stays near the designed target. Third (c), in-dataset range harmonic excitation then release, tracking MSE to the rest observation.}

Across all three tests, ABCD-based models overall drift less than their plain counterparts. This is especially the case for the Koopman and VONs which provide highly accurate predictions and reasonable extrapolations in the tested scenarios. Fig.~\ref{fig:sim_results}(b), bottom left, shows VON oscillator and stiffness forces, explaining stability under extrapolated pressures. Here, excitation forces predominantly push the SCR to the right, while stiffness forces act in opposite direction back to rest position, creating a stable equilibrium.

\section{Conclusions}
\rev{We presented the first demonstration, to our knowledge, of reliable long-horizon open-loop control of a physical soft robot using interpretable latent dynamics learned solely from video. We achieved this with Visual Oscillator Networks (VONs) and the Attention Broadcast Decoder (ABCD)~\cite{krauss2025learning}, using improved training with multi-step and rest-state losses together with implicit damping. These adjustments support single-shooting in latent space to accurately track image-specified waypoints without camera feedback. Target images for these trajectories, including deformations beyond the training pressure range, are obtained with an interactive live simulator and then mapped to model-specific latent waypoints. On the physical robot we evaluated normal and extrapolated setpoints as well as various dynamic trajectories. ABCD-based models consistently outperform their plain counterparts. Koopman with ABCD achieves the strongest overall performance, followed by the VON, consistent with~\cite{krauss2025learning}: Koopman provides more accurate long-horizon prediction, while the VON offers mechanically interpretable dynamics and empirically stable behavior, with only slightly higher tracking error. Simulation stress tests of Koopman with ABCD and the VON confirm stable static holding, accurate extrapolated equilibria, and recovery after strong harmonic excitation.}

Future work should move from planar open-loop to \rev{three-dimensional} closed or partially feedback-stabilized control, e.g. sensor and end-effector feedback in the model learning. Further, future models should account for low-level pressure controller dynamics to overcome current limitations.


\bibliographystyle{IEEEtran}
\bibliography{references}

@article{sugiyama2024latent,
  title={Latent representation-based learning controller for pneumatic and hydraulic dual actuation of pressure-driven soft actuators},
  author={Sugiyama, Taku and Kutsuzawa, Kyo and Owaki, Dai and Hayashibe, Mitsuhiro},
  journal={Soft Robotics},
  volume={11},
  number={1},
  pages={105--117},
  year={2024},
  publisher={SAGE Publications Sage CA: Los Angeles, CA}
}

@INPROCEEDINGS{tonkens2021soft,
  author={Tonkens, Sander and Lorenzetti, Joseph and Pavone, Marco},
  booktitle={2021 IEEE International Conference on Robotics and Automation (ICRA)}, 
  title={Soft Robot Optimal Control Via Reduced Order Finite Element Models}, 
  year={2021},
  volume={},
  number={},
  pages={12010-12016},
  doi={10.1109/ICRA48506.2021.9560999}}

@ARTICLE{armanini2023soft,
  author={Armanini, Costanza and Boyer, Frédéric and Mathew, Anup Teejo and Duriez, Christian and Renda, Federico},
  journal={IEEE Transactions on Robotics}, 
  title={Soft Robots Modeling: A Structured Overview}, 
  year={2023},
  volume={39},
  number={3},
  pages={1728-1748},
  doi={10.1109/TRO.2022.3231360}}

@ARTICLE{krauss2025learning,
  author={Krauss, Henrik and Licher, Johann and Takeishi, Naoya and Raatz, Annika and Yairi, Takehisa},
  journal={IEEE Robotics and Automation Letters},
  title={Learning Visually Interpretable Oscillator Networks for Soft Continuum Robots From Video},
  year={2026},
  volume={11},
  number={8},
  pages={9495-9502},
  doi={10.1109/LRA.2026.3703241}
}

@article{takeishi2017learning,
  title={Learning Koopman invariant subspaces for dynamic mode decomposition},
  author={Takeishi, Naoya and Kawahara, Yoshinobu and Yairi, Takehisa},
  journal={Advances in neural information processing systems},
  volume={30},
  year={2017}
}

@inproceedings{bartholdt_parameter_2021,
  title = {A {{Parameter Identification Method}} for {{Static Cosserat Rod Models}}: {{Application}} to {{Soft Material Actuators}} with {{Exteroceptive Sensors}}},
  shorttitle = {A {{Parameter Identification Method}} for {{Static Cosserat Rod Models}}},
  booktitle = {2021 {{IEEE}}/{{RSJ International Conference}} on {{Intelligent Robots}} and {{Systems}} ({{IROS}})},
  author = {Bartholdt, Max and Wiese, Mats and Schappler, Moritz and Spindeldreier, Svenja and Raatz, Annika},
  year = 2021,
  pages = {624--631},
  urldate = {2025-04-23}
}

@article{lusch2018deep,
  title={Deep learning for universal linear embeddings of nonlinear dynamics},
  author={Lusch, Bethany and Kutz, J Nathan and Brunton, Steven L},
  journal={Nature communications},
  volume={9},
  number={1},
  pages={4950},
  year={2018},
  publisher={Nature Publishing Group UK London}
}

@inproceedings{stoelzle2024input,
  title={Input-to-state stable coupled oscillator networks for closed-form model-based control in latent space},
  author={Stölzle, M. and Della Santina, C.},
  booktitle={Advances in Neural Information Processing Systems},
  year={2024}
}

@inproceedings{higgins2017beta,
  title={beta-vae: Learning basic visual concepts with a constrained variational framework},
  author={Higgins, Irina and Matthey, Loic and Pal, Arka and Burgess, Christopher and Glorot, Xavier and Botvinick, Matthew and Mohamed, Shakir and Lerchner, Alexander},
  booktitle={International conference on learning representations},
  year={2017}
}

@article{brunton2022modern,
author = {Brunton, Steven L. and Budi\v{s}i\'{c}, Marko and Kaiser, Eurika and Kutz, J. Nathan},
title = {Modern Koopman Theory for Dynamical Systems},
journal = {SIAM Review},
volume = {64},
number = {2},
pages = {229-340},
year = {2022},
doi = {10.1137/21M1401243},
eprint = {https://doi.org/10.1137/21M1401243}
}

@ARTICLE{alkayas2025structure,
  author={Alkayas, Abdulaziz Y. and Mathew, Anup Teejo and Feliu-Talegon, Daniel and Zweiri, Yahya and Thuruthel, Thomas George and Renda, Federico},
  journal={IEEE Robotics and Automation Letters}, 
  title={Structure-Preserving Model Order Reduction of Slender Soft Robots via Autoencoder-Parameterized Strain}, 
  year={2025},
  volume={10},
  number={10},
  pages={11006-11013},
  doi={10.1109/LRA.2025.3606389}}

@misc{licher2025adaptive,
      title={Adaptive Model-Predictive Control of a Soft Continuum Robot Using a Physics-Informed Neural Network Based on Cosserat Rod Theory}, 
      author={Johann Licher and Max Bartholdt and Henrik Krauss and Tim-Lukas Habich and Thomas Seel and Moritz Schappler},
      year={2025},
      eprint={2508.12681},
      archivePrefix={arXiv},
      primaryClass={cs.RO},
      url={https://arxiv.org/abs/2508.12681}, 
}

@article{haggerty2023control,
  title={Control of soft robots with inertial dynamics},
  author={Haggerty, David A and Banks, Michael J and Kamenar, Ervin and Cao, Alan B and Curtis, Patrick C and Mezi{\'c}, Igor and Hawkes, Elliot W},
  journal={Science robotics},
  volume={8},
  number={81},
  pages={eadd6864},
  year={2023},
  publisher={American Association for the Advancement of Science}
}

@article{almanzor_static_2023,
  title = {Static {{Shape Control}} of {{Soft Continuum Robots Using Deep Visual Inverse Kinematic Models}}},
  author = {Almanzor, Elijah and Ye, Fan and Shi, Jialei and Thuruthel, Thomas George and Wurdemann, Helge A. and Iida, Fumiya},
  year = 2023,
  journal = {IEEE Transactions on Robotics},
  volume = {39},
  number = {4},
  pages = {2973--2988},
  urldate = {2024-04-17}
}

@article{cenedese_datadriven_2022,
  title = {Data-Driven Modeling and Prediction of Non-Linearizable Dynamics via Spectral Submanifolds},
  author = {Cenedese, Mattia and Ax{\aa}s, Joar and B{\"a}uerlein, Bastian and Avila, Kerstin and Haller, George},
  year = 2022,
  journal = {Nature Communications},
  volume = {13},
  number = {1},
  pages = {872},
  publisher = {Nature Publishing Group},
  urldate = {2026-03-16},
  copyright = {2022 The Author(s)}
}

@article{thuruthel_control_2018,
  title = {Control {{Strategies}} for {{Soft Robotic Manipulators}}: {{A Survey}}},
  shorttitle = {Control {{Strategies}} for {{Soft Robotic Manipulators}}},
  author = {Thuruthel, Thomas George and Ansari, Yasmin and Falotico, Egidio and Laschi, Cecilia},
  year = 2018,
  journal = {Soft Robotics},
  volume = {5},
  number = {2},
  pages = {149--163},
  publisher = {Mary Ann Liebert, Inc., publishers},
  urldate = {2026-03-10}
}

@article{falotico_learning_2025,
  title = {Learning {{Controllers}} for {{Continuum Soft Manipulators}}: {{Impact}} of {{Modeling}} and {{Looming Challenges}}},
  shorttitle = {Learning {{Controllers}} for {{Continuum Soft Manipulators}}},
  author = {Falotico, Egidio and Donato, Enrico and Alessi, Carlo and Setti, Elisa and Nazeer, Muhammad Sunny and Agabiti, Camilla and Caradonna, Daniele and Bianchi, Diego and Piqu{\'e}, Francesco and Ansari, Yasmin Tauqeer and Killpack, Marc},
  year = 2025,
  journal = {Advanced Intelligent Systems},
  volume = {7},
  number = {2},
  pages = {2400344},
  urldate = {2025-03-26},
  copyright = {\copyright{} 2024 The Author(s). Advanced Intelligent Systems published by Wiley-VCH GmbH}
}

@article{alora_discovering_2025a,
  title = {Discovering Dominant Dynamics for Nonlinear Continuum Robot Control},
  author = {Alora, John Irvin and Cenedese, Mattia and Haller, George and Pavone, Marco},
  year = 2025,
  journal = {npj Robotics},
  volume = {3},
  number = {1},
  pages = {5},
  publisher = {Nature Publishing Group},
  urldate = {2026-03-16},
  copyright = {2025 This is a U.S. Government work and not under copyright protection in the US; foreign copyright protection may apply}
}

@article{marquesmonteiro_visuodynamic_2024,
  title = {Visuo-Dynamic Self-Modelling of Soft Robotic Systems},
  author = {Marques Monteiro, Richard and Shi, Jialei and Wurdemann, Helge and Iida, Fumiya and George Thuruthel, Thomas},
  year = 2024,
  journal = {Frontiers in Robotics and AI},
  volume = {11},
  publisher = {Frontiers},
  urldate = {2025-11-07}
}

\end{document}